\documentclass[10pt,twocolumn,letterpaper]{article}

\usepackage[pagenumbers]{iccv} 

\newcommand{\red}[1]{{\color{red}#1}}

\usepackage{graphicx}
\usepackage{amsmath}
\usepackage{amssymb}
\usepackage{booktabs}
\usepackage{multirow} 
\usepackage{graphicx} 
\usepackage{wasysym}
\usepackage{utfsym}
\usepackage{colortbl} 
\usepackage{array} 
\usepackage{marvosym}

\newcommand{\model}{\textbf{\textit{p-MoD}}}

\renewcommand{\thefootnote}{*}

\definecolor{iccvblue}{rgb}{0.21,0.49,0.74}
\usepackage[pagebackref,breaklinks,colorlinks,allcolors=iccvblue]{hyperref}

\def\paperID{} 
\def\confName{ICCV}
\def\confYear{2025}

\title{ \textit{p-MoD}: Building Mixture-of-Depths MLLMs via Progressive Ratio Decay }

\author{
Jun Zhang\textsuperscript{1,*}, Desen Meng\textsuperscript{1,*}, Zhengming Zhang \textsuperscript{2},
Zhenpeng Huang \textsuperscript{1}, Tao Wu \textsuperscript{1}, Limin Wang \textsuperscript{1,3,~\Letter}
\\
\small$^1$State Key Laboratory for Novel Software Technology, Nanjing University \quad
\small$^2$China Mobile Research Institute \quad
$^3$Shanghai AI Lab\\
\url{https://github.com/MCG-NJU/p-MoD}
}

\begin{document}
\maketitle
\begin{abstract}

Despite the remarkable performance of multimodal large language models (MLLMs) across diverse tasks, the substantial training and inference costs impede their advancement. In this paper, we propose \model{}, an efficient MLLM architecture that significantly reduces training and inference costs while maintaining model performance.
The majority of computation in MLLMs stems from the overwhelming volume of vision tokens processed by the transformer-based LLM. Accordingly, we leverage the \textbf{Mixture-of-Depths} (MoD) mechanism, where each LLM layer selects essential vision tokens to process while skipping redundant ones. However, integrating MoD into MLLMs is non-trivial. To address the challenges of training and inference stability as well as limited training data, we adapt the MoD module with two novel designs: tanh-gated weight normalization (\textbf{TanhNorm}) and symmetric token reweighting (\textbf{STRing}). Moreover, we observe that vision tokens exhibit higher redundancy in deeper layers and thus design a progressive ratio decay (\textbf{PRD}) strategy, which gradually reduces the token retention ratio layer by layer, employing a shifted cosine schedule. This crucial design fully unleashes the potential of MoD, significantly boosting the efficiency and performance of our models. 
Extensive experiments on two baseline models across 15 benchmarks show that our model matches or even surpasses the performance of corresponding baselines, while requiring only 55.6\% TFLOPs and 53.7\% KV cache storage during inference, and 77.7\% GPU hours during training.

\end{abstract}

{
\renewcommand{\thefootnote}%
{\fnsymbol{footnote}}
\footnotetext[0]{*~Equal contribution. \Letter~Corresponding author.}
}
\vspace{-10pt}
\section{Introduction}
\label{sec:intro}

\begin{figure}[tp]
    \includegraphics[width=0.9\linewidth]{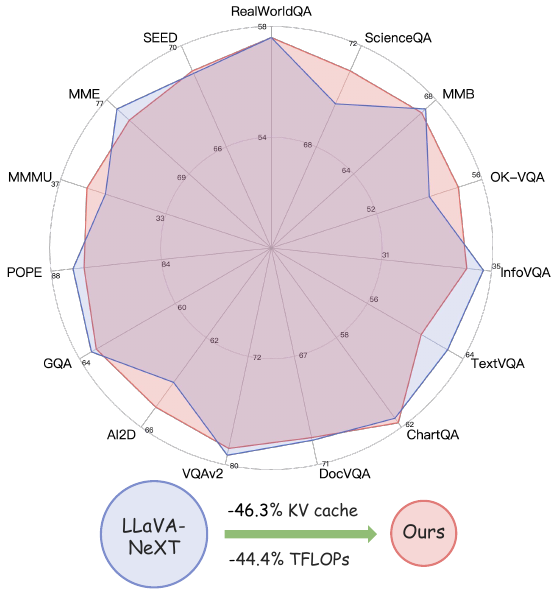}
    \vspace{0cm}
    \caption{
    \textbf{Comparison with the LLaVA-NeXT~\cite{liu2024llavanext} baseline.} Our model demonstrates comparable performance with the baseline model on 15 benchmarks across various domains, with 46.3\% fewer KV cache storage and 44.4\% fewer TFLOPs during inference. 
    }
    \label{fig:intro}
    \vspace{-0.3cm}
\end{figure}

Recently, both academia and industry have witnessed the rapid development of multimdodal large language models (MLLMs)~\cite{wang2024qwen2,li2024llava,zhang2024internlm,hong2024cogvlm2,ye2024mplug,agrawal2024pixtral,tong2024cambrian}, which have demonstrated exceptional performance across various vision-language understanding tasks. Pioneering works in this field~\cite{liu2024visual, zhu2023minigpt, liu2023improvedllava, instructblip} focused on processing single low-resolution image inputs. Subsequently, to meet the diverse demands of real-world applications, increasing efforts have broadened the forms of visual inputs that MLLMs can support, including multiple high-resolution images and videos~\cite{li2024llava, wang2024qwen2, zhang2024internlm, chen2024far,team2023gemini}.

Current state-of-the-art MLLMs handle high-resolution images either by dividing the original image into multiple slices~\cite{zhang2024internlm, liu2024llavanext, chen2024far} which are independently processed by the vision encoder, or by using a stronger vision encoder with improved positional encoding which can process any image at its native resolution~\cite{liu2024oryx,wang2024qwen2,agrawal2024pixtral}. Consequently, when processing multiple high-resolution images or videos, the number of vision tokens increases dramatically, proportional to the number of pixels and the number of images or video frames. The overwhelming volume of vision tokens processed by the transformer-based LLM results in explosion of computational costs, which severely hampers further development and broader application of MLLMs. Designing more efficient MLLM architectures with minimal performance degradation has thus become an urgent challenge for the community.

Previous efforts primarily focus on compressing vision tokens \textbf{before} the LLM, either in the Vision Encoder or the multimodal projector~\cite{chen2024far,zhang2024internlm,mckinzie2024mm1,yao2024deco,bai2023qwen, han2024free, shang2024llava, yang2025visionzip, zhang2024cls, omri2025token}. These approaches forces the LLM to handle heavily compressed vision information, overlooking the fact that the LLM, with its enormous model capacity, has the potential to compress the vision tokens by itself within the transformer layer.

In this paper, we focus on optimizing computation efficiency of MLLMs \textbf{within} the transformer layers of the LLM. We propose to build efficient MLLMs with Mixture-of-Depths (\textbf{MoD})~\cite{raposo2024mixture} mechanism, which selects the most important and informative vision tokens to be processed by each transformer layer, while skipping redundant ones to improve efficiency. However, integrating MoD mechanism into MLLMs is non-trivial and entails several significant challenges.

First, different from training an MoD-based LLM from scratch, integrating MoD mechanism into a pre-trained vanilla LLM during multimodal training poses a substantial risk of disrupting the language abilities of the original LLM. To address this, we design tanh-gated weight normalization (\textbf{TanhNorm}), which not only ensures proper initialization of the newly added MoD module, but also enhances training stability and performance. It also mitigates numerical stability issues during inference.

Second, MLLMs are trained on multimodal data~\cite{liu2024llavanext} that are several orders of magnitude smaller in scale compared to the text data used for training MoD-based LLMs~\cite{raposo2024mixture}.  We enhance the MoD mechanism with symmetric token reweighting (\textbf{STRing}) module which fully leverage the language supervision signals during training, enabling MoD modules to learn to accurately assess token importance even with limited training data.

Thanks to these two enhancements, our upgraded MoD layers can be seamlessly applied to MLLMs. However, we argue that setting a \textbf{fixed} ratio of retained tokens across different MoD layers~\cite{raposo2024mixture} is a suboptimal design choice under multimodal scenario, as the degree of redundancy in vision tokens should vary across layers. We conducted a series of exploratory experiments by adjusting the ratio of different layers in our MoD-based MLLM, which demonstrate that vision tokens exhibit \textbf{higher} redundancy in \textbf{deeper} layers. Accordingly, we propose a progressive ratio decay (\textbf{PRD}) strategy, which follows a shifted cosine schedule to gradually reduce the token retention ratio layer by layer. Trained with this strategy, our model significantly outperforms models that use a constant retention ratio across all layers under the same computation budget.

Building upon the above innovations, we present our model, \model{}, Mixture-of-Depth MLLMs equipped with our \space\textbf{\textit{p}}rogressive ratio decay strategy and upgraded MoD layers (\ie \model{} layers). Extensive experiments validate the effectiveness of our proposed model. As shown in Figure \ref{fig:intro}, across 15 benchmarks spanning various domains, our model matches or even outperforms the strong LLaVA-NeXT~\cite{liu2024llavanext} baseline, with only 55.6\% TFLOPs and 53.7\% KV cache storage during inference, and 77.7\% GPU hours during training.

\section{Related Work}
\label{sec:related}
\begin{figure*}[tp]
    \includegraphics[width=\linewidth]{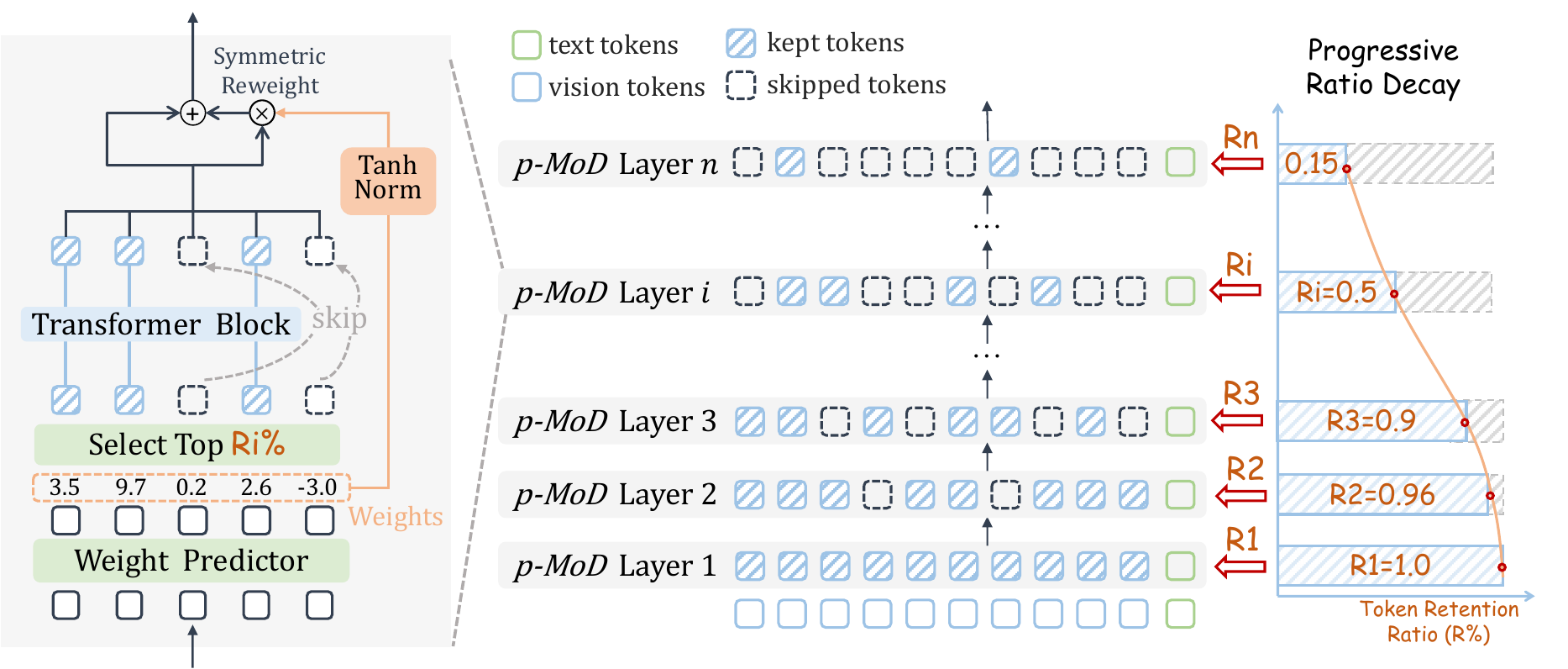}
    \vspace{-0.5cm}
    \caption{
        \textbf{Overview of \space\model{}}. \quad
        \textit{Middle}: Our efficient MLLM model which consists of \model{} layers. Each layer independently selects important vision tokens to process while skipping redundant ones. The proportion of selected tokens for layer $i$ is specified by the token retention ratio \(R_i\). For simplicity, the vision encoder and connector are omitted.
        \quad
        \textit{Left}: Detailed architecture of \model{} layers. Given the input tokens, the weight predictor first assigns weights to each token. The top $R_i$\% of tokens with highest weights are selected and processed by the transformer layer, while the rest of the tokens are skipped. The weights are normalized by \textbf{TanhNorm} module, and then both the selected and skipped tokens are \textit{symmetrically} scaled by their corresponding weights in our \textbf{STRing} module. 
        \quad
        \textit{Right}: Our crucial design is the progressive ratio decay(\textbf{PRD}) strategy, which gradually reduces the token retention ratio $R_i$ layer by layer, following a shifted cos schedule. 
    }
    \label{fig:pmod}
    \vspace{-0.3cm}
\end{figure*}

\noindent\textbf{Building Efficient LLMs.}
Considerable efforts have been made to build efficient LLMs. One representative approach is the Mixture of Experts (\textbf{MoE}) mechanism~\cite{shazeer2017outrageously,fedus2022switch,lepikhin2020gshard,zoph2022st,dai2024deepseekmoe,jiang2024mixtral}, where a router directs each token to different MLP experts of the transformer block. MoE models achieve faster training and inference speeds compared to dense models of the same size, but at the cost of lower performance. The more recently proposed Mixture of Depths (\textbf{MoD})~\cite{raposo2024mixture} module assigns weights to tokens and processes only a portion of highest-weighted tokens, skipping lower-weighted tokens to save computation. MoD-based LLMs demonstrate strong performance under limited compute budgets. In this paper, we propose to build more efficient MLLMs by upgrading MoD with several innovative improvements.
\\

\noindent\textbf{Building Efficient MLLMs.}
The main computational burden in MLLMs arises from the large number of vision tokens that the LLM must process. Previous works mainly focused on compressing vision tokens \textbf{before} they enter the LLM  via convolution~\cite{chen2024far,zhang2024internlm}, pooling~\cite{mckinzie2024mm1,yao2024deco}, query-based modules~\cite{bai2023qwen, hu2024matryoshka} or token merging~\cite{chai2024auroracap, li2024videochat, yang2025visionzip, shang2024llava, hu2024illava}. However, reducing vision-related computation \textbf{within} the LLM layers is less explored. LLaVolta~\cite{chen2024llavolta} speeds up MLLM training by performing naive average pooling operation in intermediate LLM layers. 
FastV~\cite{chen2025image} and some subsequent works \cite{he2024zipvl, zhang2024sparsevlm, huang2024mini, xing2024pyramiddrop, hu2024illava} improve MLLMs' efficiency by compressing vision tokens in intermediate transformer layers based on attention scores. However, these methods neglect the fact that tokens compressed in early layers might be crucial in deeper layers, potentially leading to performance degradation.
In contrast, we utilize MoD for learnable \textbf{layerwise} vision token \textbf{selection}, improving model efficiency while avoiding potential performance loss caused by token dropping.

\section{Method}
\label{sec:method}
In this section, we introduce our \model{} model.  As illustrated in Figure \ref{fig:pmod}, our model consists of \model{} layers which upgrades MoD architecture with tanh-gated weight normalization(\textbf{TanhNorm}) and 
symmetric token reweighting(\textbf{STRing}). Our crucial design is the progressive ratio decay(\textbf{PRD}) strategy which controls the token retention ratio across different layers. In the following sections, we first revisit MoD briefly and then explain each component of our \model{} model in detail.

\subsection{Revisiting Mixture-of-Depths}
\label{sec:MoD-revisit}
The MoD layer consists of a weight predictor and a vanilla transformer layer. It assigns weights for each input token using the weight predictor and then selects the top $R\%$ tokens with the highest weights to be processed by the transformer layer. Formally, given an MoD layer $M$ with a token retention ratio $R$, a transformer layer $T$, and a linear weight predictor, the input tokens $X \in \mathbb{R}^{n\times d}$ is first passed through the linear predictor to generate a set of weights:
\begin{align}
    w = \text{Linear}(X) \in \mathbb{R}^{n},
\end{align}
with $n$ denoting the sequence length and $d$ denoting the embedding dimension. 

Then, we compute the $R$-th percentile of the router weights, denoted as $P_R(w)$. Tokens with weights larger than $P_R(w)$ will be selected to be processed by the transformer layer $T$ while other tokens are skipped in this layer, which guarantees only $n\times R\%$ tokens are processed. The calculation process of an MoD layer can be formulated as:
\begin{align}
    X_i^{\prime} = 
    \begin{cases}
        w_iT(X_i) + T(X_i) , & \text{if} \quad w_i > P_R(w) \\
        X_i, & \text{if}\quad w_i \leq P_R(w)
    \end{cases}\space.
    \label{eq:MoD}
\end{align}
Notably, after being processed by the transformer layer, the selected tokens are then scaled by their corresponding weights. In this way, the weight predictor is engaged into the gradient path, enabling its parameters to be updated by backpropagation. We term this process \textit{token reweighting}.

In the original MoD module designed for LLMs~\cite{raposo2024mixture}, $X$ represents text tokens. In our multimodal scenario, $X$ represents vision tokens. We only apply MoD on vision tokens as they occupy the main computation load and exhibit high redundancy in LLM layers.

\subsection{Adapting Mixture-of-Depths Module}
Different from training MoD-based LLMs from scratch, integrating MoD into MLLMs presents several challenges. The insertion of new MoD modules into pre-trained LLMs can lead to instability during training and inference. The relatively small amount of multimodal data may not be sufficient to train the MoD modules. In this section, we introduce two enhancements to the MoD module to address these challenges.

\begin{figure*}[tp]
  \centering
  \begin{subfigure}[b]{0.44\textwidth}
    \includegraphics[width=\textwidth]{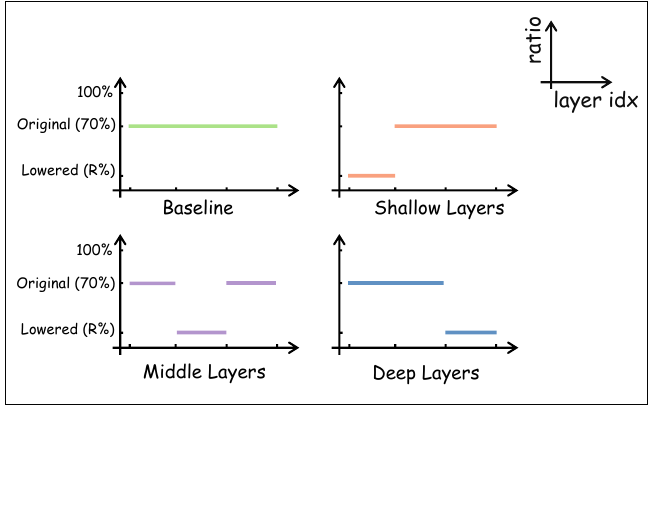}
    \caption{\textbf{Illustration of our exploratory experiment setup.} Our baseline MoD model is trained with a token retention ratio of 70\% for all layers. We divide the layers into three groups: shallow, middle, and deep. In each set of experiments, we decrease the token retention ratio (R\%) in a specific group of layers.}
    \label{fig:explore_exp-a}
  \end{subfigure}
  \hfill
  \begin{subfigure}[b]{0.54\textwidth}
    \includegraphics[width=\textwidth]{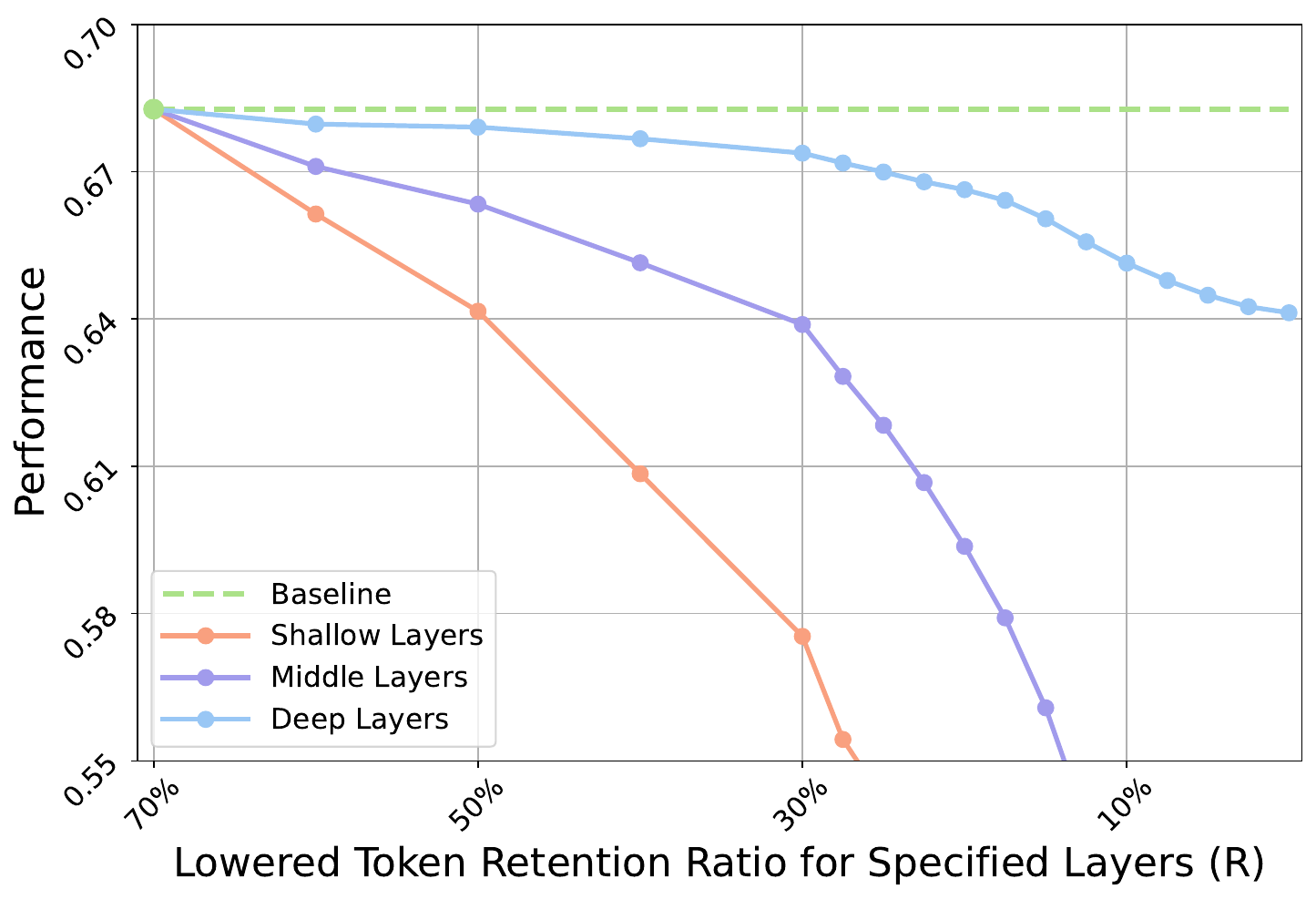}
    \caption{\textbf{Results of our exploratory experiments.} Performance drops more slowly when the token retention ratio of deeper layers is decreased, indicating that vision tokens are more \textit{redundant} in \textit{deeper} layers. This inspires us to \textit{progressively reduce} the token retention ratio layer by layer.}
      \label{fig:explore_exp-b}
  \end{subfigure}
  \caption{\textbf{Exploratory experiments on vision token redundancy.}}
  \label{explore_experiment}
  \vspace{-0.3cm}
\end{figure*}

\subsubsection{Tanh-gated Weight Normalization}
\label{sec:tanh}

The original MoD mechanism \cite{raposo2024mixture} is designed for training MoD-based LLMs from scratch. In our multimodal training scenario, we need to insert new MoD modules into a pre-trained vanilla LLM. One common approach is alternate training \cite{zhang2024beyond}, where only the newly inserted modules are trained in the first stage, and all modules are simultaneously trained in the second stage. However, this approach is not suitable for MoD, as the MoD modules scale the tokens by their weights during the token reweighting process, and the subsequent \textbf{frozen} LLM layers are unable to handle \textbf{scaled} tokens.
Our experiments show consistent results that training the MoD modules while freezing the LLM layers fails to converge. Consequently, the only feasible method is to directly address the challenge: inserting the MoD modules into the LLM and training both of them simultaneously.

To address this challenge, we design tanh-gated weight normalization (\textbf{TanhNorm}), which employs a normalization function $f(w) = \alpha \tanh(w)$ to normalize the predicted weights before token reweighting. After applying \textbf{TanhNorm} to MoD, Equation \ref{eq:MoD} can be reformulated as:

\vspace{-1em}
\begin{align}
    X_i^{\prime} = 
    \begin{cases}
        \alpha \tanh(w_i) T(X_i) & \\ \quad\quad\quad\quad+ T(X_i), & \text{if}\quad w_i > P_R(w) \\
        X_i, & \text{if}\quad w_i \leq P_R(w)
    \end{cases} 
    \quad.
    \label{eq:tanhnorm-MoD}
\end{align}

Here $\alpha$ is a hyper-parameter which controls the range of the normalized weights. Although simple, our \textbf{TanhNorm} ensures: (1) the normalized weight distribution is zero-centered, (2) the range(variance) of the weight distribution can be easily controlled by adjusting the gating factor $\alpha$. These properties offer the following benefits:

\begin{itemize}
    \item The normalized weights are closely around zero at the start of training, which ensures proper training initialization by keeping the LLM intact after inserting MoD modules.
    \item The zero-centered weight distribution reduced the risk that some tokens are repetitively scaled by positive(or negative) weights in all MoD layers. This ensures training stability and mitigates numerical stability issues during inference.
\end{itemize}
Both of the above benefits contribute to improved model performance. We validate the effectiveness of \textbf{TanhNorm} in Section \ref{sec:ablation}.

\begin{table*}[ht]
\centering
\resizebox{\textwidth}{!}{%
\begin{tabular}{c|cc|cccccccccc|c}
\toprule
Model           & \begin{tabular}[c]{@{}c@{}}Inference\\ TFLOPs $\downarrow$\end{tabular} & \begin{tabular}[c]{@{}c@{}}Inference\\ KV cache $\downarrow$\end{tabular} & \begin{tabular}[c]{@{}c@{}}Doc\\ VQA\end{tabular} & \begin{tabular}[c]{@{}c@{}}Chart\\ QA\end{tabular} & \begin{tabular}[c]{@{}c@{}}Text\\ VQA\end{tabular} & \begin{tabular}[c]{@{}c@{}}Info\\ VQA\end{tabular} & \begin{tabular}[c]{@{}c@{}}RW\\ QA\end{tabular} & \begin{tabular}[c]{@{}c@{}}G\\ QA\end{tabular} & \begin{tabular}[c]{@{}c@{}}OK\\ VQA\end{tabular} & \begin{tabular}[c]{@{}c@{}}PO\\ PE\end{tabular} & \begin{tabular}[c]{@{}c@{}}AI\\ 2D\end{tabular} & \begin{tabular}[c]{@{}c@{}}SE\\ ED\end{tabular} & AVG  \\ \midrule
LLaVA-v1.5      & 8.38                                                       & 100\%                                                        & 28.1                                              & 18.2                                               & 46.0                                               & 25.8                                               & 55.6                                            & 61.9                                           & 53.4                                             & 85.9                                            & 55.2                                            & 66.2                                            & 49.6 \\
\rowcolor[HTML]{EFEFEF} 
+\space\model{} & 4.92$\red{(-41.3\%)}$                                      & 53.7\%                                                       & 27.6                                              & 16.8                                               & 44.8                                               & 26.8                                               & 55.7                                            & 62.2                                           & 56.0                                             & 85.5                                            & 56.2                                            & 66.5                                            & 49.8 \\ \midrule
LLaVA-NeXT      & 39.46                                                      & 100\%                                                        & 70.1                                              & 61.6                                               & 62.7                                               & 34.7                                               & 57.6                                            & 63.5                                           & 54.0                                             & 87.2                                            & 64.0                                            & 68.9                                            & 62.4 \\
\rowcolor[HTML]{EFEFEF} 
+\space\model{} & 21.94$\red{(-44.4\%)}$                                     & 53.7\%                                                       & 70.0                                              & 61.8                                               & 60.5                                               & 34.1                                               & 57.6                                            & 63.3                                           & 55.1                                             & 86.8                                            & 65.1                                            & 69.0                                            & 62.3 \\ \bottomrule
\end{tabular}%
}
\caption{
    \textbf{Comparison with baseline models on 10 benchmarks.} Our models (marked in {\color{gray} gray}) achieves comparable or even better performance compared to baseline models, with only 55.6\% TFLOPs and 53.7\% KV cache storage during inference. All models are of 7B parameter scale.
}
\label{main_results_1}
\end{table*}

\begin{table*}[!tp]
\centering
    \begin{minipage}{0.44\linewidth}
    \centering
    \resizebox{\textwidth}{!}{%
    \begin{tabular}{c|ccccc|c}
    \toprule
    Model           & \begin{tabular}[c]{@{}c@{}}S\\ QA\end{tabular} & \begin{tabular}[c]{@{}c@{}}MM\\ B\end{tabular} & \begin{tabular}[c]{@{}c@{}}MM\\ MU\end{tabular} & \begin{tabular}[c]{@{}c@{}}VQA\\ v2\end{tabular} & \begin{tabular}[c]{@{}c@{}}MM\\ E\end{tabular} & AVG  \\ \midrule
    LLaVA-v1.5      & 69.7                                           & 64.1                                           & 36.6                                            & 76.6                                             & 1506.8                                         & 64.5 \\
    \rowcolor[HTML]{EFEFEF} 
    +\space\model{} & 69.3                                           & 65.4                                           & 36.3                                            & 76.9                                             & 1482.8                                         & 64.4 \\ \midrule
    LLaVA-NeXT      & 69.7                                           & 67.5                                           & 35.3                                            & 79.3                                             & 1519.3                                         & 65.6 \\
    \rowcolor[HTML]{EFEFEF} 
    +\space\model{} & 71.0                                           & 67.3                                           & 36.0                                            & 78.8                                             & 1495.5                                         & 65.6 \\ \bottomrule
    \end{tabular}%
    }
    \caption{\textbf{Comparison with baseline models on more benchmarks.} Our \model{} models (marked in {\color{gray} gray}) matches the performance of corresponding baseline models. SQA stands for ScienceQA-IMG. All models are 7B in size.
    } 
    \label{main_results_2}
    \end{minipage} 
\hfill
    \begin{minipage}{0.52\linewidth}
    \centering
    \resizebox{\textwidth}{!}{%
    \begin{tabular}{l|cccccc|c}
    \toprule
    Model                    & \begin{tabular}[c]{@{}c@{}}Doc\\ VQA\end{tabular} & \begin{tabular}[c]{@{}c@{}}Chart\\ QA\end{tabular} & \begin{tabular}[c]{@{}c@{}}Text\\ VQA\end{tabular} & \begin{tabular}[c]{@{}c@{}}G\\ QA\end{tabular} & \begin{tabular}[c]{@{}c@{}}SE\\ ED\end{tabular} & \begin{tabular}[c]{@{}c@{}}MM\\ B\end{tabular} & AVG           \\ \midrule
    vanilla MoD                & 61.1                                              & 54.0                                               & 55.8                                               & 61.8                                           & 66.9                                            & 63.1                                           & 60.4          \\
    +\space\textbf{TanhNorm} & 61.8                                              & 54.5                                               & 56.4                                               & 62.5                                           & 67.1                                            & 65.9                                           & 61.4          \\
    \quad+\space\textbf{STRing}   & 65.7                                              & 58.4                                               & 59.3                                               & 63.0                                           & 67.1                                            & 66.8                                           & 63.4          \\
    \rowcolor[HTML]{EFEFEF} 
    \quad\quad+\space\textbf{PRD}      & \textbf{70.0}                                     & \textbf{61.8}                                      & \textbf{60.5}                                      & \textbf{63.3}                                  & \textbf{69.0}                                   & \textbf{67.3}                                  & \textbf{65.3} \\ \bottomrule
    \end{tabular}%
    }
    \caption{\textbf{Ablation study on our propsed innovations.} Under fair comparison, all of our proposed modules (TanhNorm, STRing, and PRD) significantly improve model performance \textit{without} any computational overhead, making indispensable contributions to the strong performance of our final model (marked in {\color{gray} gray}).}
    \label{tab:main_ablation}
    \end{minipage}
    \label{tabel_5bench_and_ablation}
    \vspace{-0.3cm}
\end{table*}

\subsubsection{Symmetric Token Reweighting}
Compared to the text data used for training MoD-based LLMs, MLLMs are trained on multimodal data~\cite{liu2024llavanext} that are several orders of magnitude smaller in scale. It is challenging to train a weight predictor (the only learnable part in the MoD module) to accurately and robustly assess the importance of vision tokens and assign corresponding weights. 

As stated in Section \ref{sec:MoD-revisit}, the gradient of the weight predictor module stems from the token reweighting process. The original MoD performs token reweighting only on selected tokens as in Equation \ref{eq:MoD}. In this way, the language supervision signals only supervise the weight predictor to assign high weights to the selected tokens, while the process of predicting weights for the skipped tokens is not supervised (\ie no gradient).  

To fully leverage the limited training data, we enhance the MoD mechanism by symmetrically applying the token reweighting process to both selected and skipped tokens. With our symmetric token reweighting (\textbf{STRing}) module, Equation \ref{eq:tanhnorm-MoD} can be further modified as:
\begin{align}
    X_i^{\prime} = 
    \begin{cases}
        \alpha \tanh(w_i) T(X_i) & \\ 
        \quad\quad\quad\quad+ T(X_i), & \text{if} \quad w_i > P_R(w) \\
        \alpha \tanh(w_i)X_i + X_i, & \text{if}\quad w_i \leq P_R(w)
    \end{cases}
    \quad.
    \label{eq:string-MoD}
\end{align}

In this way, the MoD module can fully leverage the language supervision signals during training and learn to accurately assess token importance with limited training data.

\subsection{Progressive Ratio Decay}
\label{sec:PRD}

After upgrading MoD with \textbf{TanhNorm} and \textbf{STRing} modules, we are able to successfully train MoD-based MLLMs. However, the performance-efficiency trade-off exhibited by the model is far from satisfactory. When the token retention ratio is set below 70\%, the model's performance deteriorates sharply. 

Original MoD-based LLMs~\cite{raposo2024mixture} adopt a fixed token retention ratio across all MoD layers. This strategy assumes that the tokens have the same degree of redundancy in different layers. We argue that this assumption does not hold in the multimodal scenario. Under the effect of self-attention in every transformer layer, vision tokens gradually aggregate information from each other and text tokens gather relevant information from vision tokens. Therefore, vision tokens are expected to become increasingly redundant in deeper layers.

To validate our hypothesis, we conduct a series of exploratory experiments on our MoD-based MLLM trained with a token retention ratio of 70\% across all layers. As shown in Figure \ref{fig:explore_exp-a}, we first divide the layers into several groups: shallow, middle, and deep layers. In each set of experiment, we gradually decrease the token retention ratio (R\%) in a specific group of layers while keeping the other layers unchanged, and evaluate the model under this customized inference setting. The results in Figure \ref{fig:explore_exp-b} strongly support our hypothesis: performance drops more \textit{slowly} when
the token retention ratio of \textit{deeper} layers is decreased.
This suggests that vision tokens exhibit higher redundancy in deeper layers, which is consistent with observations in previous works \cite{chen2025image, zhang2024sparsevlm}.

Accordingly, we design a progressive ratio decay (PRD) strategy. As shown on the right side of Figure \ref{fig:pmod}, \textbf{PRD} gradually reduces the token retention ratio layer by layer, following a \textbf{shifted cosine schedule}. Suppose the model has $L$ layers, the token retention ratio for the $l$-th layer is formulated as:
\vspace{-1em}
\begin{align}
    R_l = \cfrac{1}{2}\cos{\cfrac{\pi l}{L}} + \beta, \quad l = 1,2,...,L
    \label{eq:PRD}.
\end{align}
Here $\beta$ is a \textbf{shift factor}, which can be used to flexibly control the overall computational cost of the model by vertically shifting the cosine decay curve.

Experiments in Section \ref{sec:ablation} demonstrates that our \model{} model with \textbf{PRD} strategy significantly outperforms models that use a constant retention ratio under the same computation budget. It also outperforms other kinds of ratio schedulers.

\begin{figure*}[!tp]
\begin{minipage}{0.63\linewidth}
\centering
\renewcommand\arraystretch{1.35}
\resizebox{1.0\linewidth}{!}{
\begin{tabular}{cc|cc|cccccc|c}
\toprule
\multicolumn{2}{c|}{\begin{tabular}[c]{@{}c@{}}Ratio\\ Scheduler\end{tabular}} & Decay                           & Progressive                     & \begin{tabular}[c]{@{}c@{}}Doc\\ VQA\end{tabular} & \begin{tabular}[c]{@{}c@{}}Chart\\ QA\end{tabular} & \begin{tabular}[c]{@{}c@{}}Text\\ VQA\end{tabular} & \begin{tabular}[c]{@{}c@{}}G\\ QA\end{tabular} & \begin{tabular}[c]{@{}c@{}}SE\\ ED\end{tabular} & \begin{tabular}[c]{@{}c@{}}MM\\ B\end{tabular} & AVG                                   \\ \midrule
\multicolumn{2}{c|}{Constant}                                                  & $\usym{2717}$                   & $\usym{2717}$                   & 58.0                                              & 51.6                                               & 56.1                                               & 62.1                                           & 64.5                                            & 62.8                                           & 59.2                                  \\
\multicolumn{2}{c|}{Interleaved}                                               & $\usym{2717}$                   & $\usym{2717}$                   & 60.3                                              & 54.8                                               & 57.0                                               & 62.3                                           & 65.3                                            & 62.1                                           & 60.3                                  \\
\multicolumn{2}{c|}{Stepped}                                                   & $\usym{2714}$                   & $\usym{2717}$                   & 67.2                                              & 61.2                                               & 59.4                                               & 63.5                                           & 68.4                                            & 66.1                                           & 64.3                                  \\ \midrule
                                                   & Linear                    &                                 &                                 & 69.3                                              & 60.4                                               & \textbf{60.5}                                      & \textbf{63.8}                                  & 68.0                                            & 66.7                                           & 64.8                                  \\ \cmidrule(lr){2-2}
\multirow{-2}{*}{\textbf{PRD}}                     & Cosine                    & \multirow{-2}{*}{$\usym{2714}$} & \multirow{-2}{*}{$\usym{2714}$} & \cellcolor[HTML]{EFEFEF}\textbf{70.0}             & \cellcolor[HTML]{EFEFEF}\textbf{61.8}              & \cellcolor[HTML]{EFEFEF}\textbf{60.5}              & \cellcolor[HTML]{EFEFEF}63.3                   & \cellcolor[HTML]{EFEFEF}\textbf{69.0}           & \cellcolor[HTML]{EFEFEF}\textbf{67.3}          & \cellcolor[HTML]{EFEFEF}\textbf{65.3} \\ \bottomrule
\end{tabular}
}
\captionof{table}{
\textbf{
    Ablation on different token retention ratio schedules.
} Our default model is marked in {\color{gray} gray}.
}
\label{table:ablation_4funcs}
\end{minipage} 
\hspace{0.25cm}
\begin{minipage}{0.3\linewidth}
  \centering
   \includegraphics[width=0.9\linewidth]{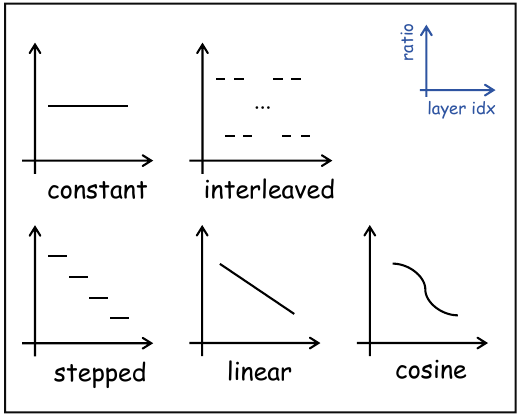}
   \caption{\textbf{Illustration of different ratio schedule functions in Table \ref{table:ablation_4funcs}.}}
   \label{fig:ablation_4funcs}
\end{minipage}
\end{figure*}

\begin{table*}[!ht]
\centering
\resizebox{\textwidth}{!}{%
\begin{tabular}{l|cc|cccccccccccc|c}
\toprule
Model                             & \begin{tabular}[c]{@{}c@{}}Token Compression \\ Ratio $\downarrow$\end{tabular} & \begin{tabular}[c]{@{}c@{}}Infer.\\ TFLOPs\end{tabular} & \begin{tabular}[c]{@{}c@{}}Doc\\ VQA\end{tabular} & \begin{tabular}[c]{@{}c@{}}Chart\\ QA\end{tabular} & \begin{tabular}[c]{@{}c@{}}Text\\ VQA\end{tabular} & \begin{tabular}[c]{@{}c@{}}Info\\ VQA\end{tabular} & \begin{tabular}[c]{@{}c@{}}RW\\ QA\end{tabular} & GQA                         & SQA                         & \begin{tabular}[c]{@{}c@{}}MM\\ MU\end{tabular} & \begin{tabular}[c]{@{}c@{}}PO\\ PE\end{tabular} & \begin{tabular}[c]{@{}c@{}}AI\\ 2D\end{tabular} & \begin{tabular}[c]{@{}c@{}}VQA\\ v2\end{tabular} & \begin{tabular}[c]{@{}c@{}}SE\\ ED\end{tabular} & AVG                         \\ \midrule
{\color[HTML]{9B9B9B} LLaVA-NeXT} & 100\%                                                                            & {\color[HTML]{9B9B9B} 39.46}                            & {\color[HTML]{9B9B9B} 70.1}                       & {\color[HTML]{9B9B9B} 61.6}                        & {\color[HTML]{9B9B9B} 62.7}                        & {\color[HTML]{9B9B9B} 34.7}                        & {\color[HTML]{9B9B9B} 57.6}                     & {\color[HTML]{9B9B9B} 63.5} & {\color[HTML]{9B9B9B} 69.7} & {\color[HTML]{9B9B9B} 35.3}                     & {\color[HTML]{9B9B9B} 87.2}                     & {\color[HTML]{9B9B9B} 64.0}                     & {\color[HTML]{9B9B9B} 79.3}                      & {\color[HTML]{9B9B9B} 68.9}                     & {\color[HTML]{9B9B9B} 62.9} \\
$\quad$+$\ $MQT~\cite{hu2024matryoshka}                  & 50.2\%                                                                           & 19.86                                                   & 49.9                                              & 44.0                                               & 53.5                                               & 29.8                                               & 53.5                                            & 62.3                        & 69.1                        & 36.1                                            & 85.8                                            & 64.0                                            & 77.2                                             & 65.8                                            & 57.6                        \\
$\quad$+$\ $LLaVolta~\cite{chen2024llavolta}             & 53.1\%                                                                           & 21.30                                                   & 66.6                                              & 58.8                                               & 60.1                                               & 33.2                                               & 56.9                                            & \textbf{63.7}               & 70.2                        & \textbf{36.7}                                   & 86.3                                            & \textbf{65.3}                                   & 78.6                                             & 68.5                                            & 62.0                        \\
$\quad$+$\ $FastV~\cite{chen2025image}                 & 53.1\%                                                                           & 22.73                                                   & 65.9                                              & 58.2                                               & \textbf{62.2}                                      & 33.0                                               & 55.6                                            & 62.4                        & 69.9                        & 35.6                                            & 84.5                                            & 63.6                                            & 78.7                                             & 68.2                                            & 61.5                        \\
\rowcolor[HTML]{EFEFEF} 
$\quad$+$\ $$\model{}$            & 53.7\%                                                                           & 21.94                                                   & \textbf{70.0}                                     & \textbf{61.8}                                      & 60.5                                               & \textbf{34.1}                                      & \textbf{57.6}                                   & 63.3                        & \textbf{71.0}               & 36.0                                            & \textbf{86.8}                                   & 65.1                                            & \textbf{78.8}                                    & \textbf{69.0}                                   & \textbf{62.8}               \\ \bottomrule
\end{tabular}%
}
\caption{\textbf{Comparison with other vision token compression methods.} \model{} significantly outperforms other compression methods, achieving the best performance on most benchmarks and the best average performance. All models are of 7B parameter scale. Results on 13B models are demonstrated in Table \ref{tab:13B_comp}.}
\label{tab:compare_with_others}
\vspace{-0.3cm}
\end{table*}

\section{Experiment}
\subsection{Setups}
\label{sec:setups}

\textbf{Models.}
To evaluate the effectiveness of \model{}, we select two representative open-source MLLMs: LLaVA-1.5~\cite{liu2023improvedllava} and LLaVA-NeXT~\cite{liu2024llavanext}, as the baseline models in our experiments. LLaVA-1.5 resizes input images to the fixed resolution which aligns with the training setup of its CLIP~\cite{CLIP-radford2021learning} vision encoder, encoding an image into 576 tokens. LLaVA-NeXT divides high-resolution images into multiple slices which are independently processed by the vision encoder. This strategy enhances the visual perception capabilities, but also leads to a significantly larger number of vision tokens (up to 2880) and much higher computation costs. 

\noindent \textbf{Training Recipe.}
Our \model{} models follow the two-stage training recipe of the baselines. In the pre-training stage, the MLP connector is trained on image caption data. In the fine-tuning stage, \model{} modules are insert into the LLM and updated together with the connector and the LLM.

\noindent \textbf{Benchmarks.}
We conduct comprehensive experiments across 15 benchmarks: VQAv2~\cite{goyal2017making}, GQA~\cite{hudson2019gqa}, OK-VQA~\cite{okvqa-marino2019ok}, AI2D~\cite{ai2d-kembhavi2016diagram} and ScienceQA-IMG~\cite{scienceqa-lu2022learn} are traditional visual question answering benchmarks;  DocVQA~\cite{mathew2021docvqa}, TextVQA~\cite{textvqa-singh2019towards}, ChartQA~\cite{masry2022chartqa} and InfographicVQA~\cite{infovqa-mathew2022infographicvqa} focus on fine-grained visual question answering; POPE~\cite{POPE-li2023evaluating} evaluates hallucination in MLLMs;  SEED-Bench (Image)~\cite{li2023seed-bench}, RealWorldQA, MME~\cite{fu2024mmecomprehensiveevaluationbenchmark}, MMBench~\cite{liu2025mmbench} and MMMU~ \cite{yue2024mmmu} are comprehensive benchmarks tailored for MLLMs. To ensure our results can be conveniently reproduced, we evaluate our model on all these benchmarks with the lmms-eval~\cite{zhang2024lmms} evaluation framework. 

\noindent \textbf{Evaluation Metrics.} In addition to performance on benchmarks, we present various metrics that reflect training and inference efficiency. We report TFLOPs which measures the inference computation complexity, and the KV cache storage which constitutes the main memory bottleneck during inference. Furthermore, we measure GPU hours consumed during training, along with inference latency.

\subsection{Main Results}

We integrate \model{} with LLaVA-1.5 and LLaVA-NeXT baseline models and conduct comprehensive evaluation across 15 benchmarks, which measures comprehensive abilities of MLLMs across various aspects. The results are shown in Table \ref{main_results_1} and Table \ref{main_results_2}. Both \model{}-LLaVA-v1.5 and \model{}-LLaVA-NeXT achieve comparable or even better performance across all benchmarks compared to their baseline models, with significant savings in inference TFLOPs and inference KV cache savings.

In Table \ref{main_results_1}, it is noteworthy that text-rich and graph-based benchmarks like DocVQA, ChartQA, TextVQA, and InfoVQA require fine-grained visual perception and reasoning abilities. Models can only given a correct answer when they successfully identify answer-related regions that occupy only small portions of the entire image. Remarkably, \model{}-LLaVA-NeXT achieves negligible performance drop on DocVQA and InfoVQA, and even gains 0.2\% accuracy on ChartQA with substantial improvements in time and memory efficiency.

The above results indicate that our approach substantially improves efficiency while maintaining performance. 

\subsection{Ablation Study}
\label{sec:ablation}

\noindent \textbf{Effectiveness of propsed innovations.} We conduct ablation in Table \ref{tab:main_ablation} to demonstrate the effectiveness of our proposed innovations. All experiments in the table are conducted under the \textit{same} computational cost. TanhNorm, STRing and PRD \textit{all} significantly enhance model performance, making indispensable contributions to the strong performance of our final \model{} model.

\vspace{0.2cm}
\noindent \textbf{Progressive Ratio Decay.} Based on the exploratory experiments in Section \ref{sec:PRD}, we conclude that vision tokens exhibit higher redundancy in deeper layers, and design the progressive ratio decay strategy with shifted cosine schedule. 

To further validate this conclusion and the effectiveness of \textbf{PRD}, we experiment with four other schedule functions which control the token retention ratio for each LLM layer. As illustrated in Figure \ref{fig:ablation_4funcs}, the functions include: (1) a constant function that uses the same ratio for all layers, (2) an interleaved function which interleaves a vanilla transformer layer and an MoD layer with low ratio, which is adopted in the original MoD paper~\cite{raposo2024mixture}, (3) a stepped decay function, (4) a linear decay function. To ensure a fair comparison,  we fix the average retention ratios across all layers at approximately 54\%. Based on the results demonstrated in Table \ref{table:ablation_4funcs}, we can draw the following conclusions:
\begin{itemize}
    \item The non-decaying functions (\ie constant and interleaved) significantly underperform the decaying functions. This result strongly supports our conclusion that vision tokens exhibit higher redundancy in deeper layers.
    \item Our progressively decayed \textbf{PRD} functions (\ie cosine and linear) notably outperforms the discontinuous stepped decay function, and the cosine function achieves the best results. This proves the effectiveness of our shifted cosine schedule. 
\end{itemize}

\noindent \textbf{Tanh-gated Weight Normalization.} To validate the effectiveness of \textbf{TanhNorm}, we compare different weight normalization methods in Table \ref{table:ablation_allmod}. First, we verify that using no weight normalization module will result in overflow during training, as shown in the first row (details are explained in appendix). Then, we validate the effectiveness of the two important properties of \textbf{TanhNorm} stated in Section \ref{sec:tanh}: (1) the normalized weight distribution is zero-centered, (2) the range(variance) of the weight distribution can be easily controlled by adjusting the gating factor $\alpha$.

For the first property, we design two experiments with the Softmax function, a standard normalization function that is \textit{not} zero-centered. Specifically, we experiment with the vanilla softmax function $f(w) = \alpha \cdot Softmax(w), \alpha = 0.2$, and a shifted softmax function $f(w) = \alpha \cdot Softmax(w) + b, \alpha=0.4, b=-0.2$ which has the same range as our \textbf{TanhNorm} function $f(w) = \alpha \tanh(w), \alpha = 0.2$. 

Comparing the last row of Table \ref{table:ablation_allmod} with the second and third rows, we verify that \textbf{TanhNorm} significantly outperforms the vanilla softmax function and the shifted softmax function. It proves that being zero-centered is key to \textbf{TanhNorm}'s strong performance. The shifted softmax function, despite having the same range as \textbf{TanhNorm}, results in worse performance due to the lack of zero-centered property.

For the second property, we experiment with a naive choice of the gating factor $\alpha$ by setting it to 1, which can be interpreted as no gating is applied.
As shown in the fourth row of Table \ref{table:ablation_allmod}, this experiment results in the overflow problem during inference. This validates the importance of controlling the gating factor $\alpha$ to guarantee training and inference stability.

\vspace{0.2cm}

\begin{table}[tp]
\centering
\setlength\tabcolsep{4.0pt}
\resizebox{1.0\linewidth}{!}{
\begin{tabular}{c|cccccc|c}
\toprule
\begin{tabular}[c]{@{}c@{}}Norm\\ Type\end{tabular} & \begin{tabular}[c]{@{}c@{}}Doc\\ VQA\end{tabular} & \begin{tabular}[c]{@{}c@{}}Chart\\ QA\end{tabular} & \begin{tabular}[c]{@{}c@{}}Text\\ VQA\end{tabular} & \begin{tabular}[c]{@{}c@{}}G\\ QA\end{tabular} & \begin{tabular}[c]{@{}c@{}}SE\\ ED\end{tabular} & \begin{tabular}[c]{@{}c@{}}MM\\ B\end{tabular} & AVG           \\ \midrule
-                                                   & \multicolumn{6}{c|}{OVERFLOW}                                                                                                                                                                                                                                                                                   & -             \\ \midrule
Softmax                                             & 63.9                                              & 57.6                                               & 57.9                                               & 62.7                                           & 68.2                                            & 67.0                                           & 62.9          \\
Shifted softmax                                     & 62.5                                              & 57.0                                               & 57.3                                               & 62.9                                           & 68.3                                            & 66.4                                           & 62.4          \\
\textbf{TanhNorm$(\alpha=1)$}                       & \multicolumn{6}{c|}{OVERFLOW}                                                                                                                                                                                                                                                                                   & -             \\ \midrule
\rowcolor[HTML]{EFEFEF} 
\textbf{TanhNorm$(\alpha=0.2)$}                     & \textbf{70.0}                                     & \textbf{61.8}                                      & \textbf{60.5}                                      & \textbf{63.3}                                  & \textbf{69.0}                                   & \textbf{67.3}                                  & \textbf{65.3} \\ \bottomrule
\end{tabular}
}
\caption{
\textbf{Ablation on tanh-gated weight normalization}. Our default model is marked in gray. The rows with no performance reported indicates that the corresponding experiment faces overflow issue during training or inference.}
\label{table:ablation_allmod}
\vspace{-0.4cm}
\end{table}

\subsection{Comparison with Related Works}
In Table \ref{tab:compare_with_others}, we compare \model{} with several strong token compression methods. MQT~\cite{hu2024matryoshka} employs a query transformer to compress vision tokens. LLaVolta~\cite{chen2024llavolta} performs average pooling in intermediate LLM layers, achieving remarkable results despite its simplicity. FastV~\cite{chen2025image} drops vision tokens in intermediate LLM layers based on the attention score from text tokens. \model{} significantly outperforms other compression methods, achieving the best performance on most benchmarks and the best average performance.

\begin{table*}[!ht]
\centering
\resizebox{0.9\linewidth}{!}{
\begin{tabular}{c|cccc|ccccccc}
\toprule
                                     & \multicolumn{4}{c|}{Effiency}                                                                                                                                                                                                                                                                                            & \multicolumn{7}{c}{Benchmark}                                                                                                                                                                                                                                                                                                                                      \\ \cmidrule(l){2-12} 
\multirow{-2}{*}{Model}              & \begin{tabular}[c]{@{}c@{}}Training\\ GPU hours $\downarrow$\end{tabular} & \begin{tabular}[c]{@{}c@{}}Inference\\ TFLOPs $\downarrow$\end{tabular} & \begin{tabular}[c]{@{}c@{}}Inference\\ Latency (ms)  $\downarrow$\end{tabular} & \begin{tabular}[c]{@{}c@{}}Inference\\ KV cache storage $\downarrow$\end{tabular} & \begin{tabular}[c]{@{}c@{}}Doc\\ VQA\end{tabular} & \begin{tabular}[c]{@{}c@{}}Chart\\ QA\end{tabular} & \begin{tabular}[c]{@{}c@{}}Text\\ VQA\end{tabular} & \begin{tabular}[c]{@{}c@{}}G\\ QA\end{tabular} & \begin{tabular}[c]{@{}c@{}}SE\\ ED\end{tabular} & \multicolumn{1}{c|}{\begin{tabular}[c]{@{}c@{}}MM\\ B\end{tabular}} & AVG                         \\ \midrule
{\color[HTML]{9B9B9B} LLaVA-NeXT-7B} & {\color[HTML]{9B9B9B} 560}                                                & {\color[HTML]{9B9B9B} 39.46}                                            & {\color[HTML]{9B9B9B} 519.1}                                                    & {\color[HTML]{9B9B9B} 100\%}                                                      & {\color[HTML]{9B9B9B} 70.1}                       & {\color[HTML]{9B9B9B} 61.6}                        & {\color[HTML]{9B9B9B} 62.7}                        & {\color[HTML]{9B9B9B} 63.5}                    & {\color[HTML]{9B9B9B} 68.9}                     & \multicolumn{1}{c|}{{\color[HTML]{9B9B9B} 67.5}}                    & {\color[HTML]{9B9B9B} 65.7} \\
$\model{}$-0.3                       & 403 $\red{(-28.0\%)}$                                                     & 17.78 $\red{(-54.9\%)}$                                                 & 326.7 $\red{(-37.1\%)}$                                                         & 42.3\%                                                                            & 65.5                                              & 57.8                                               & 58.1                                               & 63.1                                           & 68.6                                            & \multicolumn{1}{c|}{67.0}                                           & 63.4                        \\
$\model{}$-0.4                       & 415 $\red{(-25.9\%)}$                                                     & 19.56 $\red{(-50.4\%)}$                                                 & 347.1 $\red{(-33.1\%)}$                                                         & 47.5\%                                                                            & 66.5                                              & 59.6                                               & 58.4                                               & 63.8                                           & 68.5                                            & \multicolumn{1}{c|}{66.6}                                           & 63.9                        \\
\rowcolor[HTML]{EFEFEF} 
$\model{}$-0.5                       & 435 $\red{(-22.3\%)}$                                                     & 21.94 $\red{(-44.4\%)}$                                                 & 368.0 $\red{(-29.1\%)}$                                                         & 53.7\%                                                                            & 70.0                                              & 61.8                                               & 60.5                                               & 63.3                                           & 69.0                                            & \multicolumn{1}{c|}{\cellcolor[HTML]{EFEFEF}67.3}                   & 65.3                        \\ \bottomrule
\end{tabular}
}
\caption{\textbf{The Efficiency-Performance trade-off experiments.} To showcase the versatility of our approach, we conducted experiments with $\beta= 0.3, 0.4, 0.5$, respectively. Inference latency is measured on TextVQA dataset.
}
\label{tab:tradeoff}
\vspace{0cm}
\end{table*}

\begin{figure*}[ht]
    \centering
    \includegraphics[width=0.89\linewidth]{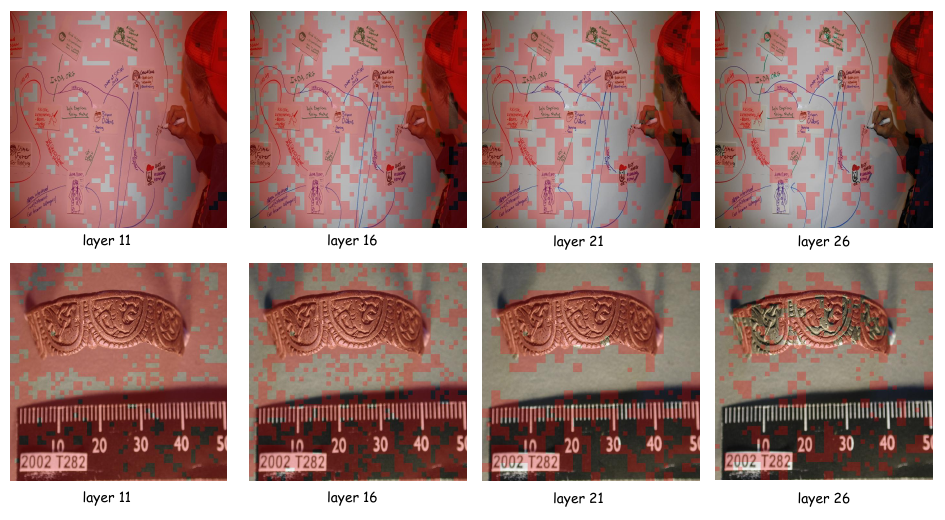}
    \caption{\textbf{Visualization of tokens selected by different \model{} layers.} The selected tokens are colored in \red{red}. Please zoom in for a clearer view. As the number of selected tokens gradually decreases in deeper layers (due to \textbf{PRD} strategy), the model gradually concentrates on vision tokens corresponding to regions with \textbf{rich semantic information}.}
    \label{vis}
\end{figure*}

\subsection{Efficiency-Performance Trade-off}
\label{sec:trade-off}
 Thanks to the shifted cosine schedule used by PRD in Section \ref{sec:PRD} to control the token retention ratio across all layers, we can change the shift factor $\beta$ to control the overall computational cost of the model. In this way, we can train and deploy \model{} models under different performance-efficiency trade-offs based on our actual needs.
 
To showcase the versatility of our approach, we conducted experiments with $\beta= 0.3, 0.4, 0.5$, respectively. The results are presented in Table \ref{tab:tradeoff}. Our default \model{} model with $\beta=0.5$ reduced training GPU hours by 22.3\%, inference TFLOPs by 44.4\%, inference latency by 29.1\%, and KV cache storage by 46.3\%. Using a lower $\beta$ can further improve efficiency, with an acceptable performance drop. 

\subsection{Visualization: Which tokens are selected?}

In Figure \ref{vis}, we visualize the tokens selected by \model{}-LLaVA-NeXT at different layers. Guided by the \textbf{PRD} strategy, the number of tokens processed by the MoD layers gradually decreases layer by layer. It can be observed that the selected vision tokens progressively concentrate on key regions in the image with \textbf{rich semantic information}. 
For the upper row of images in Figure \ref{vis}, the selected tokens gradually converge to those corresponding to text, drawings, and the person on the right. Tokens corresponding to the white background are skipped in deep layers. In the lower row of images in Figure \ref{vis}, the selected tokens gradually converge towards the artifact and the measurement markings on the ruler. These results indicate that our model effectively compresses visual information by selecting the informative tokens, enabling it to maintain performance while significantly reducing training and inference costs.

\section{Conclusion and Future Work}

In this paper, we explore building efficient MLLMs by adapting the Mixture-of-Depths mechanism. Our 
 proposed model, termed \model{}, features three key designs: tanh-gated weight normalization, symmetric token reweighting module, and the progressive ratio decay strategy. Our model achieves comparable or even superior results to the baseline models on a diverse set of 15 benchmarks, with substantial improvements on training and inference efficiency. We hope \model{} can serve as a strong and efficient baseline for future research on developing efficient MLLMs.

\clearpage
\noindent \textbf{Acknowledgements.} This work is supported by the National Key R$\&$D Program of China (No. 2022ZD0160900), Jiangsu Frontier Technology Research and Development Program (No. BF2024076), the Collaborative Innovation Center of Novel Software Technology and Industrialization, and Nanjing University-China Mobile Communications Group Co., Ltd. Joint Institute.

{
    \small

    \bibliographystyle{ieeenat_fullname}
}

\clearpage

\twocolumn[{%
\renewcommand\twocolumn[1][]{#1}%
\maketitlesupplementary
\appendix
}]

\noindent This supplementary material includes the following sections:
\begin{itemize}
    \item In Section \ref{sec:more_exp}, we provide more experiment results.
    \item In Section \ref{sec:vis_all_layers}, we show visualization results of \model{} token selection decisions across different layers.
    \item In Section \ref{sec:more_discuss}, we make some further discussions about our work.
    \item In Section \ref{sec:implementation_details}, we provide more implementation details.
\end{itemize}

\section{More Experiments}
\label{sec:more_exp}
\subsection{Experiments on Larger Model Size}
\label{sec:13B}

In Table \ref{tab:13B_comp}, we compare our method against other token compression methods on LLaVA-NeXT-13B baseline model. The results demonstrate that our method significantly outperforms other methods on larger model size, validating the scalability of our approach. 

\subsection{Experiments on Visual Grounding}

In Table \ref{tab:refcoco}, we compare \model{} with other token compression methods on visual grounding benchmarks RefCOCO, RefCOCO+ and RefCOCOg~\cite{kazemzadeh2014referitgame,yu2016modeling}. Our method significantly outperforms other methods, but it still exhibits some performance drop compared to the baseline model. This suggests that current token pruning methods still exhibit limitations on visual grounding, which might be one of the major challenges that future works in this field should aim to address.

\subsection{Compairson with More Relate Works}
In addition to comparing \model{} against three strong token compression methods in Table \ref{tab:compare_with_others}, we provide comparison with more methods in Table \ref{tab:compare_more_methods}.  
Under the same token compression ratio, \model{} demonstrates the best overall performance on our comprehensive evaluation suite covering 15 benchmarks.

\subsection{Comparison with $\gamma$-MoD}
\label{sec:gamma_comparison}

Concurrent to our work, $\gamma$-MoD~\cite{luo2024gamma} also propose to integrate Mixture-of-Depths mechanism into MLLMs. In this section, we first analyze the difference between our approach and theirs. Then we conduct experiments to show that our \model{} approach outperforms $\gamma$-MoD.

The core design of $\gamma$-MoD is computing attention map (ARank) on some samples to identify which layers MoD can be applied to. In contrast, \model{} can be effectively applied to \textbf{every} layer thanks to our TanhNorm and STRing modules. Furthermore, we propose \textbf{PRD} strategy to progressively reduce the token retention ratio layer by layer, which significantly boosts performance and efficiency. 

In Table \ref{table:cmp_with_gammamode}, we compare our method with $\gamma$-MoD on LLaVA-v1.5-7B baseline. $\gamma$-MoD utilizes higher token retention ratio than \model{} (60+\% vs 53\%), but \model{} still outperforms it by a large margin. Note that we are unable to compare both methods on LLaVA-NeXT, as $\gamma$-MoD's code implementation does not support LLaVA-NeXT.

\begin{table}[t]
\centering
\resizebox{\linewidth}{!}{%
\begin{tabular}{l|cccc|c}
\toprule
\rowcolor[HTML]{FFFFFF} 
\multicolumn{1}{c|}{\cellcolor[HTML]{FFFFFF}Method} & DocVQA                      & ChartQA                     & SEED                        & GQA                         & \begin{tabular}[c]{@{}c@{}}AVG over \\ \textcolor{blue}{\textbf{15}} Tasks*\end{tabular} \\ \midrule
\rowcolor[HTML]{FFFFFF} 
{\color[HTML]{C0C0C0} \quad LLaVA-NeXT-13B}         & {\color[HTML]{C0C0C0} 73.5} & {\color[HTML]{C0C0C0} 67.2} & {\color[HTML]{C0C0C0} 71.4} & {\color[HTML]{C0C0C0} 65.2} & {\color[HTML]{C0C0C0} 66.5}                                                              \\
\rowcolor[HTML]{FFFFFF} 
\quad+~MQT                                          & 58.6                        & 54.6                        & 68.8                        & 64.2                        & 63.0                                                                                     \\
\rowcolor[HTML]{FFFFFF} 
\quad+~FastV                                        & 70.2                        & 64.7                        & 70.8                        & 64.7                        & 65.7                                                                                     \\
\rowcolor[HTML]{FFFFFF} 
\quad+~LLaVolta                                     & 70.0                        & 61.7                        & 70.5                        & 64.7                        & 65.0                                                                                     \\
\rowcolor[HTML]{EFEFEF} 
\quad+~\model{}                                     & \textbf{72.3}               & \textbf{66.2}               & \textbf{71.6}               & \textbf{65.0}               & \textbf{66.0}                                                                            \\ \bottomrule
\end{tabular}%
}
\caption{\textbf{Experiments on LLaVA-NeXT-13B}. Our method significantly outperforms other methods on larger baseline model, validating the scalability of our approach. *Average is computed on all \textcolor{blue}{\textbf{15}} benchmarks used in Table \ref{main_results_1} and \ref{main_results_2}.}
\label{tab:13B_comp}
\end{table}

\begin{table}[t]
\centering
\resizebox{\linewidth}{!}{%
\begin{tabular}{l|ccc}
\toprule
\rowcolor[HTML]{FFFFFF} 
\multicolumn{1}{c|}{\cellcolor[HTML]{FFFFFF}Method}                               & {\color[HTML]{333333} RefCOCO Val}    & RefCOCO+ Val                          & RefCOCOg Val                          \\ \midrule
\rowcolor[HTML]{FFFFFF} 
\multicolumn{1}{c|}{\cellcolor[HTML]{FFFFFF}{\color[HTML]{C0C0C0} LLaVA-NeXT-7B}} & {\color[HTML]{C0C0C0} 82.67}          & {\color[HTML]{C0C0C0} 73.73}          & {\color[HTML]{C0C0C0} 79.41}          \\
\rowcolor[HTML]{FFFFFF} 
+MQT                                                                              & {\color[HTML]{333333} 68.77}          & 59.07                                 & 63.52                                 \\
\rowcolor[HTML]{FFFFFF} 
+LLaVolta                                                                         & 79.96                                 & 70.89                                 & 76.79                                 \\
\rowcolor[HTML]{FFFFFF} 
+FastV                                                                            & 73.83                                 & 64.90                                 & 69.78                                 \\
\rowcolor[HTML]{EFEFEF} 
+\model{}                                                                         & {\color[HTML]{333333} \textbf{80.23}} & {\color[HTML]{333333} \textbf{70.94}} & {\color[HTML]{333333} \textbf{76.96}} \\ \bottomrule
\end{tabular}%
}
\caption{
\textbf{Experiments on Visual Grounding}. Our method significantly outperforms other methods on visual grounding benchmarks RefCOCO, RefCOCO+ and RefCOCOg. We report ACC@0.5 metric on the validation sets of these benchmarks.
}
\vspace{-0.8em}
\label{tab:refcoco}
\end{table}
\begin{table}[t]
\centering
\resizebox{\linewidth}{!}{%
\begin{tabular}{l|cccc|c}
\toprule
\multicolumn{1}{c|}{Model}           & DocVQA                      & ChartQA                     & SEED                        & GQA                         & \begin{tabular}[c]{@{}c@{}}AVG over\\ \textcolor{blue}{\textbf{15}} tasks*\end{tabular} \\ \midrule
{\color[HTML]{C0C0C0} LLaVA-NeXT-7B} & {\color[HTML]{C0C0C0} 70.1} & {\color[HTML]{C0C0C0} 61.6} & {\color[HTML]{C0C0C0} 68.9} & {\color[HTML]{C0C0C0} 63.5} & {\color[HTML]{C0C0C0} 63.5}                                     \\ \midrule
+~SparseVLM~\cite{zhang2024sparsevlm}                     & 67.2                        & 52.8                        & 68.1                        & 62.6                        & 62.3                                                            \\
+~VisionZip~\cite{yang2025visionzip}                     & 65.5                        & 50.3                        & 67.4                        & 61.1                        & 61.2                                                            \\
+~PyramidDrop~\cite{xing2024pyramiddrop}                   & 66.5                        & 53.3                        & 67.5                        & 61.9                        & 61.9                                                            \\
+~FasterVLM~\cite{zhang2024cls}                   & 65.9                        & 47.7                        & 68.0                        & 61.4                        & 61.4                                                            \\
+~FreeVideoLLM~\cite{han2024free}                  & 55.5                        & 36.0                        & 68.9                        & 59.3                        & 57.6                                                            \\
+~iLLaVA~\cite{hu2024illava}                        & 64.5                        & 56.6                        & 65.5                        & 61.5                        & 59.3                                                            \\
+~LLaVA-PruMerge~\cite{shang2024llava}                & 58.0                        & 44.6                        & 67.4                        & 61.2                        & 60.0                                                            \\
\rowcolor[HTML]{EFEFEF} 
+~\model{}                      & \textbf{70.0}               & \textbf{61.8}               & \textbf{69.0}               & \textbf{63.3}               & \textbf{63.4}                                                   \\ \bottomrule
\end{tabular}%
}
\caption{
\textbf{Comparison with more vision token compression methods.}
In addition to the comparison in Table \ref{tab:compare_with_others}, we make a fair comparison between \model{} and more vision token compression methods by controlling the average token retention ratio. Our methods achieves the best overall performance across 15 benchmarks.
*Average is computed on all \textcolor{blue}{\textbf{15}} benchmarks used in Table \ref{main_results_1} and \ref{main_results_2}.
}
\label{tab:compare_more_methods}
\end{table}
\begin{table*}[!ht]
\centering
\resizebox{\textwidth}{!}{
\begin{tabular}{c|c|ccccccccccccccc|c}
\toprule
Model                             & \begin{tabular}[c]{@{}c@{}}Keep\\ Ratio$\downarrow$\end{tabular} & \begin{tabular}[c]{@{}c@{}}Doc\\ VQA\end{tabular} & \begin{tabular}[c]{@{}c@{}}Info\\ VQA\end{tabular} & \begin{tabular}[c]{@{}c@{}}Chart\\ QA\end{tabular} & \begin{tabular}[c]{@{}c@{}}Text\\ VQA\end{tabular} & \begin{tabular}[c]{@{}c@{}}RW\\ QA\end{tabular} & \begin{tabular}[c]{@{}c@{}}SE\\ ED\end{tabular} & \begin{tabular}[c]{@{}c@{}}PO\\ PE\end{tabular} & \begin{tabular}[c]{@{}c@{}}MM\\ MU\end{tabular} & \begin{tabular}[c]{@{}c@{}}AI\\ 2D\end{tabular} & \begin{tabular}[c]{@{}c@{}}VQA\\ v2\end{tabular} & \begin{tabular}[c]{@{}c@{}}OK\\ VQA\end{tabular} & \begin{tabular}[c]{@{}c@{}}MM\\ E\end{tabular} & \begin{tabular}[c]{@{}c@{}}G\\ QA\end{tabular} & \begin{tabular}[c]{@{}c@{}}S\\ QA\end{tabular} & \begin{tabular}[c]{@{}c@{}}MM\\ B\end{tabular} & AVG                         \\ \midrule
{\color[HTML]{9B9B9B} LLaVA-v1.5} & {\color[HTML]{9B9B9B} 100\%}                                     & {\color[HTML]{9B9B9B} 28.1}                       & {\color[HTML]{9B9B9B} 25.8}                        & {\color[HTML]{9B9B9B} 18.2}                        & {\color[HTML]{9B9B9B} 46.0}                        & {\color[HTML]{9B9B9B} 55.6}                     & {\color[HTML]{9B9B9B} 66.2}                     & {\color[HTML]{9B9B9B} 85.9}                     & {\color[HTML]{9B9B9B} 36.6}                     & {\color[HTML]{9B9B9B} 55.2}                     & {\color[HTML]{9B9B9B} 76.6}                      & {\color[HTML]{9B9B9B} 53.4}                      & {\color[HTML]{9B9B9B} 1506.8}                  & {\color[HTML]{9B9B9B} 61.9}                    & {\color[HTML]{9B9B9B} 69.7}                    & {\color[HTML]{9B9B9B} 64.1}                    & {\color[HTML]{9B9B9B} 54.6} \\
+$\ \gamma$-MoD                   & \textgreater{}$\ $60\%                                           & 20.4                                              & 21.5                                               & \textbf{18.1}                                      & \textbf{47.5}                                      & 53.7                                            & 66.3                                            & \textbf{86.6}                                   & 35.0                                            & 55.1                                            & \textbf{77.1}                                    & 51.3                                             & 1,377.0                                        & 62.1                                           & 67.2                                           & 59.9                                           & 52.7                        \\
\rowcolor[HTML]{EFEFEF} 
+$\ \space\model{}$               & \textbf{53.7\%}                                                  & \textbf{27.6}                                     & \textbf{26.8}                                      & 16.8                                               & 44.8                                               & \textbf{55.7}                                   & \textbf{66.5}                                   & 85.5                                            & \textbf{36.3}                                   & \textbf{56.2}                                   & 76.9                                             & \textbf{56.0}                                    & \textbf{1482.8}                                & \textbf{62.2}                                  & \textbf{69.3}                                  & \textbf{65.4}                                  & \textbf{54.7}               \\ \bottomrule
\end{tabular}

}
\captionof{table}{
\textbf{Comparison with concurrent work $\gamma$-MoD. All models are of 7B parameter scale.} 
}
\label{table:cmp_with_gammamode}
\end{table*}

\begin{figure*}[ht]
    \centering
    \includegraphics[width=0.85\linewidth]{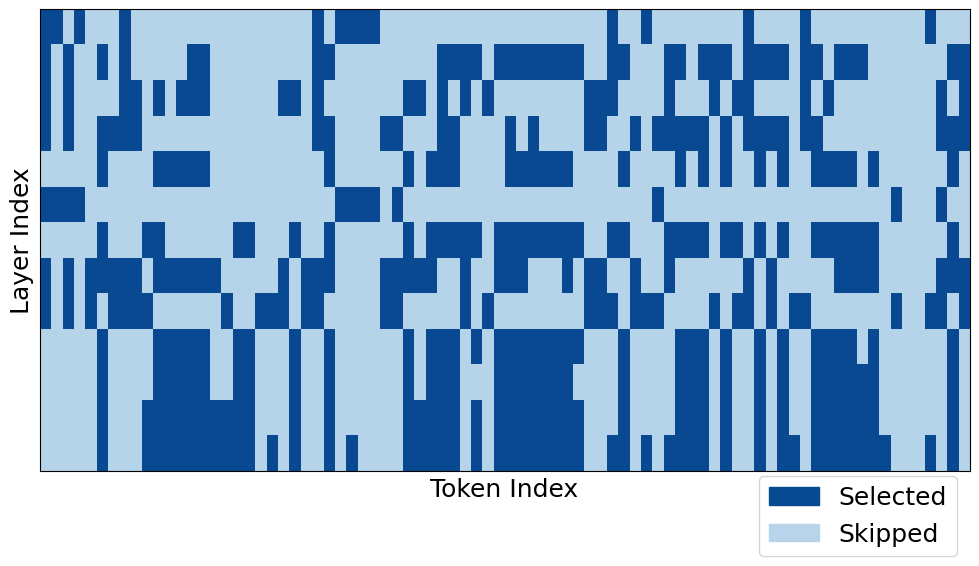}
    \caption{
    \textbf{Visualization of token selection decisions across different \model{} layers.} The horizontal axis denotes the token indexes, and the vertical axis denotes the layer indexes. It can be observed that every \model{} layer independently selects important and informative tokens.
    } 
    \label{vis_all_layers}
\end{figure*}

\begin{table*}[]
\centering
\resizebox{0.85\textwidth}{!}{%
\begin{tabular}{@{}c|c|c|c|c|c@{}}
\toprule
Name &
  Resolution &
  \begin{tabular}[c]{@{}c@{}}Train\\ Stage\end{tabular} &
  \begin{tabular}[c]{@{}c@{}}Trainable Module\&\\ Learning Rate\end{tabular} &
  Data &
  \begin{tabular}[c]{@{}c@{}}Batch\\ Size\end{tabular} \\ \midrule
\multirow{2}{*}{\model{}-LLaVA-v1.5} &
  \multirow{2}{*}{(336,336)} &
  PT &
  Connector: 1e-3 &
  558K &
  256 \\ \cmidrule(l){3-6} 
 &
   &
  SFT &
  Connector+LLM+MoD: 2e-5 &
  665K &
  128 \\ \midrule
\multirow{2}{*}{\model{}-LLaVA-NeXT} &
  \multirow{2}{*}{\begin{tabular}[c]{@{}c@{}}336 $\times$ {[}(2,2), \\ (1,2), (2,1),\\  (1,3), (3,1){]}\end{tabular}} &
  PT &
  Connector: 1e-3 &
  558K &
  256 \\ \cmidrule(l){3-6} 
 &
   &
  SFT &
  \begin{tabular}[c]{@{}c@{}}ViT: 2e-6 (baseline)\\ Connector+LLM+MoD: 2e-5\end{tabular} &
  779K &
  128 \\ \bottomrule
\end{tabular}%
}
\caption{\textbf{Detailed training configuration}. PT stands for pre-training. SFT stands for supervised fine-tuning. During fine-tuning, the vision encoder of the baseline LLaVA-NeXT model is updated, consistent with the original LLaVA-NeXT model\cite{liu2024llavanext}. We freeze the vision encoder of our \model{}-LLaVA-NeXT model for all experiments to reduce training cost.}
\label{tab:train_config}
\end{table*}

\section{Visualization of Token Selection Decisions}
\label{sec:vis_all_layers}

Figure \ref{vis_all_layers} visualizes the token selection decisions of \model{} across different layers. The horizontal axis denotes the token indexes, and the vertical axis denotes the layer indexes. It can be observed that every layer select different tokens to process, and every token is selected by different \model{} layers. This demonstrates that every \model{} layer independently selects important and informative tokens, without degrading into selecting a same set of tokens across different layers, which is identical to dropping tokens instead of layer-wise selection.

\section{Further Discussions}
\label{sec:more_discuss}

\subsection{Discussion on OCR Performance}
In our experiments, we found that \textit{all} token compression methods inevitably cause a significant performance drop on OCR-related benchmarks compared to other benchmarks. The main challenge we aim to address throughout the progress of this work is to mitigate this issue as much as possible. Experiment results in this paper show that our method achieves the
\textit{least} performance drop on OCR tasks compared to others.

\subsection{Explanation on Training Overflow Problem}
\label{sec:overflow}
In Section \ref{sec:tanh}, we mention that \textbf{TanhNorm} ensures training and inference stability. Acoordingly, we observe overflow issues in the ablation study on \textbf{TanhNorm} in Table \ref{eq:tanhnorm-MoD}. In this Section, we give a detailed explanation on the causes of overflow.

As illustrated in Equation \ref{eq:MoD}, scaling a token by excessively large weights $w_i$ across multiple LLM layers may cause \textbf{floating-point overflow}. 
For TanhNorm with $\alpha=1$ (Table \ref{eq:tanhnorm-MoD} Row 4), large $w_i$ makes $X_i^{\prime} = \alpha \tanh(w_i) T(X_i) + X_i$ approximate scaling the token $X_i$ by 2. Repeating this for the same token across multiple layers causes overflow. The same problem arises for vanilla MoD (Table \ref{eq:tanhnorm-MoD} Row 1). In this case, \textbf{no normalization} is applied to constrain the range of the weights, making it easier to produce extreme values and result in numerical overflow.

\subsection{Limitations}
One limitation of our work is that \model{} is only experimented on LLaVA-1.5 and LLaVA-NeXT models, which focus on single-image understanding tasks. We believe that our approach has the potential to achieve more remarkable results when applied on tasks that handle a larger number of vision tokens, such as multi-image and long video understanding. We leave the exploration of \model{} on other vision tasks to future research.

Another limitation is that \model{} is a \textit{trainable} method, designed for training a new MLLM from scratch with reduced training and inference costs. When being applied to trained MLLMs, it requires continual fine-tuning the model. It is not suitable for training-free scenarios.

\section{More Implementation Details}
\label{sec:implementation_details}

\subsection{Detailed Training Recipe}

Our detailed training recipe of \model{} models is shown in Table \ref{tab:train_config}. LLaVA-1.5 employs a fixed input resolution of 336$\times$336, while LLaVA-NeXT supports a pre-defined set of different resolutions (up to 672$\times$672). Both the LLaVA-v1.5 models and the LLaVA-NeXT models go through the same pre-training stage, where the MLP connector module is trained on 558K image caption data~\cite{liu2024visual} with a learning rate of 1e-3 and a batch size of 256. 

During supervised fine-tuning, LLaVA-1.5 is trained on 665K instruction-tuning data~\cite{liu2023improvedllava}, while LLaVA-NeXT is trained on a larger set of 779K data\footnote{https://huggingface.co/datasets/lmms-lab/LLaVA-NeXT-Data}. The vision encoder of the LLaVA-NeXT baseline model is updated for fine-tuning. For our \model{}-LLaVA-NeXT model, we freeze the vision encoder to save training time, as the available GPU resources are limited. When measuring the training GPU hours reported in Table \ref{tab:tradeoff}, we freeze the vision encoder for both baseline and \model{} models to ensure fair comparison.

\subsection{Hardware and Hyperparameters}
We train all our models on 8 NVIDIA RTX A6000 GPUs. The inference efficiency metrics reported in Section \ref{sec:trade-off} are measured on a single A6000 GPU. By default, we set the gating factor $\alpha$ in \textbf{TanhNorm} to 0.2, and the shift factor $\beta$ in \textbf{PRD} to $0.5$.

Due to computational constraints, we mainly use 7B models in our experiments. Experiments on 13B models are show in Table \ref{tab:13B_comp} and Section \ref{sec:13B}.

\subsection{Details on the PRD schedule.}

We find it challenging for MoD layers to learn to predict meaningful weights when the token retention ratio is set to an extremely large or small value. To address this issue, we constrain the range of the token retention ratio $R_l$ within a predefined maximum and minimum threshold.
If $R_l$ exceeds the maximum threshold, we set the token retention ratio to $1$ so that the MoD module is not applied to the $l$-th layer. If $R_l$ falls below the minimum threshold, the token retention ratio for the $l$-th layer is set to the minimum threshold:
\begin{align}
    R_l^{\prime} = 
    \begin{cases}
        1, & \text{if}\quad R_l \geq max \\
        R_l, & \text{if}\quad min < R_l < max \\
        min, & \text{if}\quad R_l \leq min
    \end{cases}
    \quad .
    \label{eq:prd-shift}
\end{align}


\end{document}